\RequirePackage[T1]{fontenc}
\documentclass[letterpaper,10pt,conference]{ieeeconf}
\IEEEoverridecommandlockouts
\usepackage{iftex}
\ifPDFTeX\usepackage[utf8]{inputenc}\fi
\usepackage[T1]{fontenc}
\usepackage{mathptmx}
\usepackage{amsmath,amsfonts}
\usepackage{booktabs,tabularx,array}
\usepackage{graphicx}
\usepackage{microtype,xcolor}
\usepackage{booktabs}
\usepackage{multirow}
\usepackage{float}

\makeatletter
\let\NAT@parse\undefined
\makeatother
\usepackage[numbers,sort&compress]{natbib}
\usepackage{xurl}
\usepackage[hidelinks]{hyperref}
\newcolumntype{Y}{>{\raggedright\arraybackslash}X}
\newcommand{\projectname}{HEARTH}

\title{\LARGE\bf \projectname{}: An Object-Centric RGB--Thermal--3D Dataset for Temperature-Aware Robot Manipulation}
\author{Yuning Su, BoRui Li, Yonghao Shi, Bofei Liu, and Xing-Dong Yang\thanks{All authors are with Simon Fraser University.}}
\hypersetup{pdfauthor={Yuning Su, BoRui Li, Yonghao Shi, Bofei Liu, Xing-Dong Yang}}

\makeatletter
\IEEEaftertitletext{\begin{minipage}{\textwidth}
\centering
\includegraphics[width=0.86\linewidth]{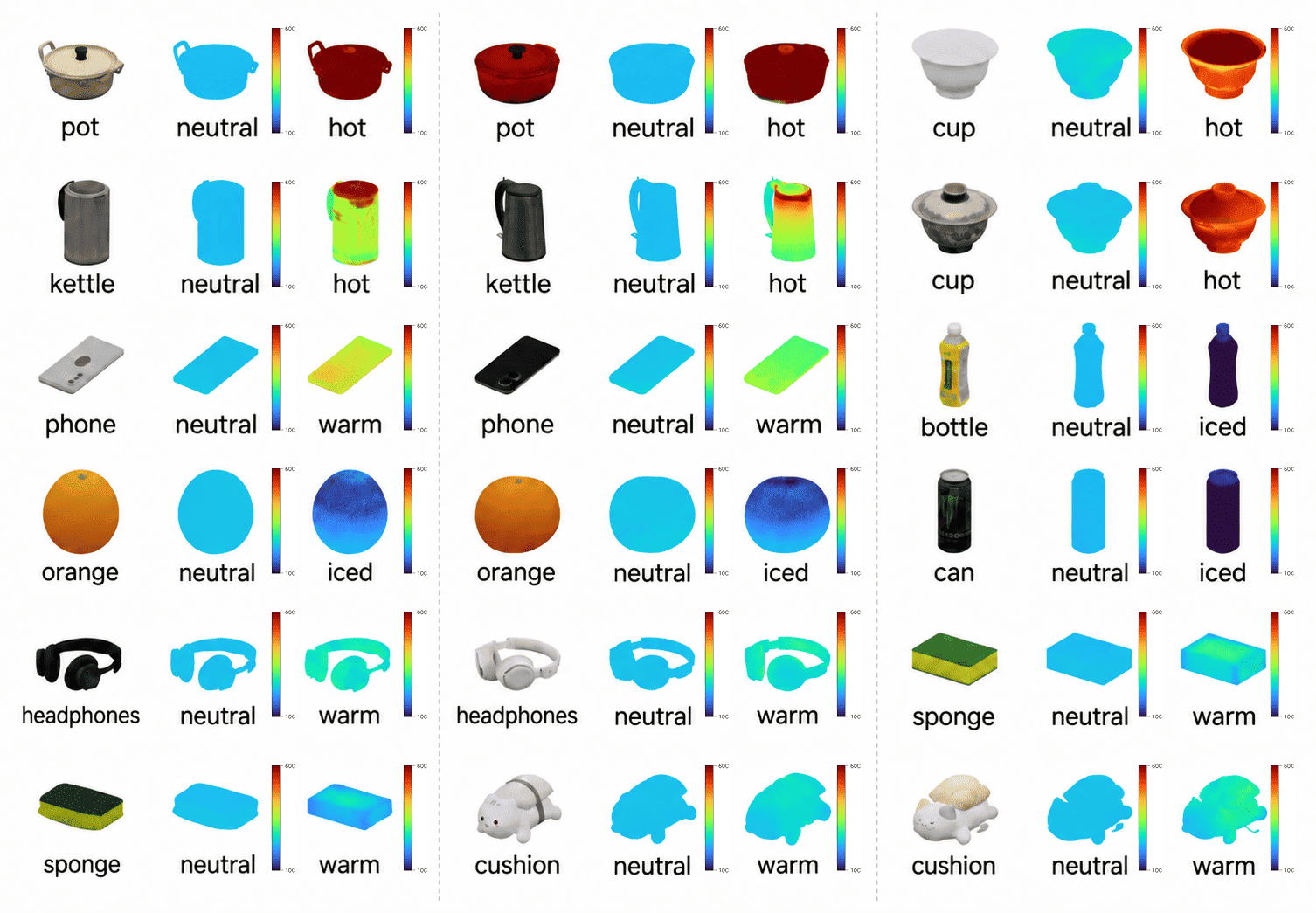}
\def\@captype{figure}
\caption{RGB appearance and two thermal states of 18 selected objects in \projectname{}. Each group shows the RGB appearance, the room-temperature reference (``neutral''), and the second captured state (hot, warm, or iced), as labeled under each panel. All thermal textures use the same Turbo color scale from 10 to 60$^\circ$C, shown beside each thermal panel.}
\label{fig:thermal_states}
\end{minipage}
}
\makeatother

\begin{document}
\maketitle
\thispagestyle{empty}
\pagestyle{empty}

\begin{abstract}
Language-guided manipulation can depend on physical properties that visible appearance does not reveal. Temperature is one such property, but object datasets for robot learning rarely associate measured temperatures with object appearance and geometry. We present \projectname{}, an object-centric RGB--thermal--3D dataset of 90 physical objects from 18 everyday categories, comprising 145 captured object states. Our pipeline maps apparent surface temperatures onto reconstructed meshes through camera calibration and pose transfer. The dataset includes raw temperature measurements, camera parameters, RGB-textured meshes, and thermal textures for simulation. We use these assets to construct three LIBERO-derived tasks and collect 1,200 demonstrations for fine-tuning a pretrained vision-language-action (VLA) model, $\pi_{0.5}$. In an ablation study, adding thermal observations to the VLA increases success on temperature-dependent object-selection tasks from 35.0\% for the RGB-only baseline to 75.0\%. These results demonstrate the utility of HEARTH for training robot policies to follow temperature-related instructions.
\end{abstract}

\section{Introduction}

Language-guided robot manipulation requires identifying both the requested object and the properties specified by the instruction. Some of these properties are not apparent from visible appearance: two identical cups can contain water at different temperatures, yet an instruction such as ``bring me the warm cup'' requires distinguishing them. Existing datasets describe object appearance, geometry, poses, and semantic categories, or provide language-annotated robot trajectories~\citep{calli2017ycb,hodan2018bop,mo2019partnet,walke2023bridgedata,openx2024}. Others include acoustic and tactile observations~\citep{gao2022objectfolder2,gao2023objectfolderbenchmark}. Object-level temperature measurements aligned with appearance and geometry remain less widely available. This gap limits the data available for learning temperature-dependent object selection.

Temperature varies both across object states and over an object's surface. A cup can retain the same appearance as its contents cool, and its handle can be cooler than its body. These variations call for spatial measurements associated with individual object instances, beyond category labels or object-level physical-property annotations~\citep{cao2025physx,Genesis}. Existing RGB--thermal datasets and thermal reconstruction methods primarily address scene perception or reconstruction~\citep{hwang2015multispectral,ha2017mfnet,jia2021llvip,hassan2025thermonerf,liu2025thermalgs}. Manipulation simulation additionally requires reusable object assets whose geometry, appearance, and thermal states can be rendered together in different scenes.

We introduce \projectname{}, an object-centric dataset pairing measured apparent surface temperatures with object appearance and reconstructed geometry. It contains 90 physical instances from 18 categories, with five instances per category. All instances have a room-temperature capture, and 55 have an additional cooled, heated, or post-use capture, giving 145 object states in total. Figure~\ref{fig:thermal_states} illustrates the variation across states and surface regions. The dataset retains the raw measurements alongside the derived thermal assets.

Constructing these assets requires aligning thermal measurements with reconstructed surfaces despite differences in image texture between the cameras. We use a rigid RGB--thermal rig and a calibration target detectable in both modalities. Camera poses estimated from the RGB views are transferred to the thermal camera using the calibrated relative pose and then manually refined. We project the thermal measurements onto the reconstructed meshes to produce thermal textures aligned with object geometry.

We evaluate the assets in three LIBERO-derived manipulation tasks using 1,200 demonstrations. Two tasks require selecting between visually identical objects in different thermal states, and the third requires selecting between two object categories with the same thermal-state label. We fine-tune two copies of the same pretrained $\pi_{0.5}$ LIBERO policy~\citep{pi05} under matched settings, with and without an additional thermal view. Across 60 evaluation episodes per policy, thermal input increases overall success from 56.7\% to 81.7\%, with a 40.0-percentage-point gain on each thermal-state selection task. Success on the category-selection task remains high at 95.0\%, compared with 100.0\% for the RGB-only baseline. This comparison evaluates temperature-related instruction following and category selection within the constructed simulation tasks.

Our contributions are:
\begin{itemize}
\item An object-centric RGB--thermal--3D dataset of 90 physical instances across 18 categories and 145 captured states, with raw apparent-temperature measurements and derived assets to be released upon acceptance.

\item A pipeline for mapping measured temperatures onto reconstructed object surfaces using RGB--thermal calibration, camera-pose transfer, and per-view refinement.

\item Three LIBERO-derived tasks and 1,200 demonstrations, with a matched comparison of $\pi_{0.5}$ policies showing improved thermal-state selection and high success on category selection with thermal input.
\end{itemize}

\section{Related Work}

\subsection{Object and Physical-Property Datasets}

Object datasets commonly provide appearance, geometry, and semantic structure. YCB includes household objects, RGB-D scans, geometric models, and metadata~\citep{calli2017ycb}. BOP supports 6D object-pose estimation~\citep{hodan2018bop}, and PartNet provides part annotations~\citep{mo2019partnet}. \projectname{} complements these resources with apparent surface temperatures registered to individual object meshes.

Datasets also capture physical properties and sensory responses. ObjectFolder~2.0 provides simulated visual, acoustic, and tactile observations~\citep{gao2022objectfolder2}, while ObjectFolder Real includes meshes, videos, impact sounds, and tactile recordings of physical objects~\citep{gao2023objectfolderbenchmark}. PhysX-3D adds annotated or generated physical properties to 3D assets~\citep{cao2025physx}. \projectname{} focuses on spatial temperature measurements captured from real-world objects in different states and registered to each object's reconstructed geometry.

\subsection{RGB--Thermal Data and Thermal 3D Reconstruction}

Paired visible and infrared datasets support perception under challenging imaging conditions. KAIST Multispectral pairs color and thermal images for pedestrian detection~\citep{hwang2015multispectral}, MFNet addresses RGB--thermal semantic segmentation for driving scenes~\citep{ha2017mfnet}, and LLVIP provides visible--infrared pairs for low-light vision~\citep{jia2021llvip}. These datasets are organized primarily around images and scenes, rather than reusable manipulation assets.

Thermal reconstruction methods recover spatial thermal representations from images. Thermal-NeRF reconstructs scenes from infrared images~\citep{ye2024thermalnerf}. ThermoNeRF jointly models RGB and thermal views and introduces ThermoScenes, which includes building facades and scenes with everyday objects~\citep{hassan2025thermonerf}. ThermalGS models thermal changes in building scenes using Gaussian splatting and introduces TSDN~\citep{liu2025thermalgs}. Our dataset instead organizes measured thermal states around individual reconstructed objects, so that the same assets can be reused across simulation scenes.

\subsection{Multimodal Robot Learning}

Large robot datasets and pretrained policies enable learning from images, language, and actions~\citep{openx2024,pi05}. ThermoAct~\citep{son2026thermoact} uses thermal observations for everyday robot tasks, combining a vision-language model (VLM) for instruction interpretation and planning with a VLA for control. Its real-world experiments include temperature-related instructions such as bringing warm water or a cold drink.

Studying such tasks in simulation requires object assets that associate measured thermal states with appearance and geometry. \projectname{} provides these assets. We evaluate their use by fine-tuning a pretrained VLA with rendered thermal observations and comparing it against an RGB-only baseline under matched training and evaluation settings.

\section{Data Acquisition}
\label{sec:acquisition}

We collect paired RGB images and temperature measurements of each object, reconstruct its geometry from the RGB views, and calibrate the two cameras so that thermal observations can be mapped onto the reconstructed surface.

\subsection{Object Selection and Thermal States}

\begin{figure}[t]
\centering
\includegraphics[width=\linewidth]{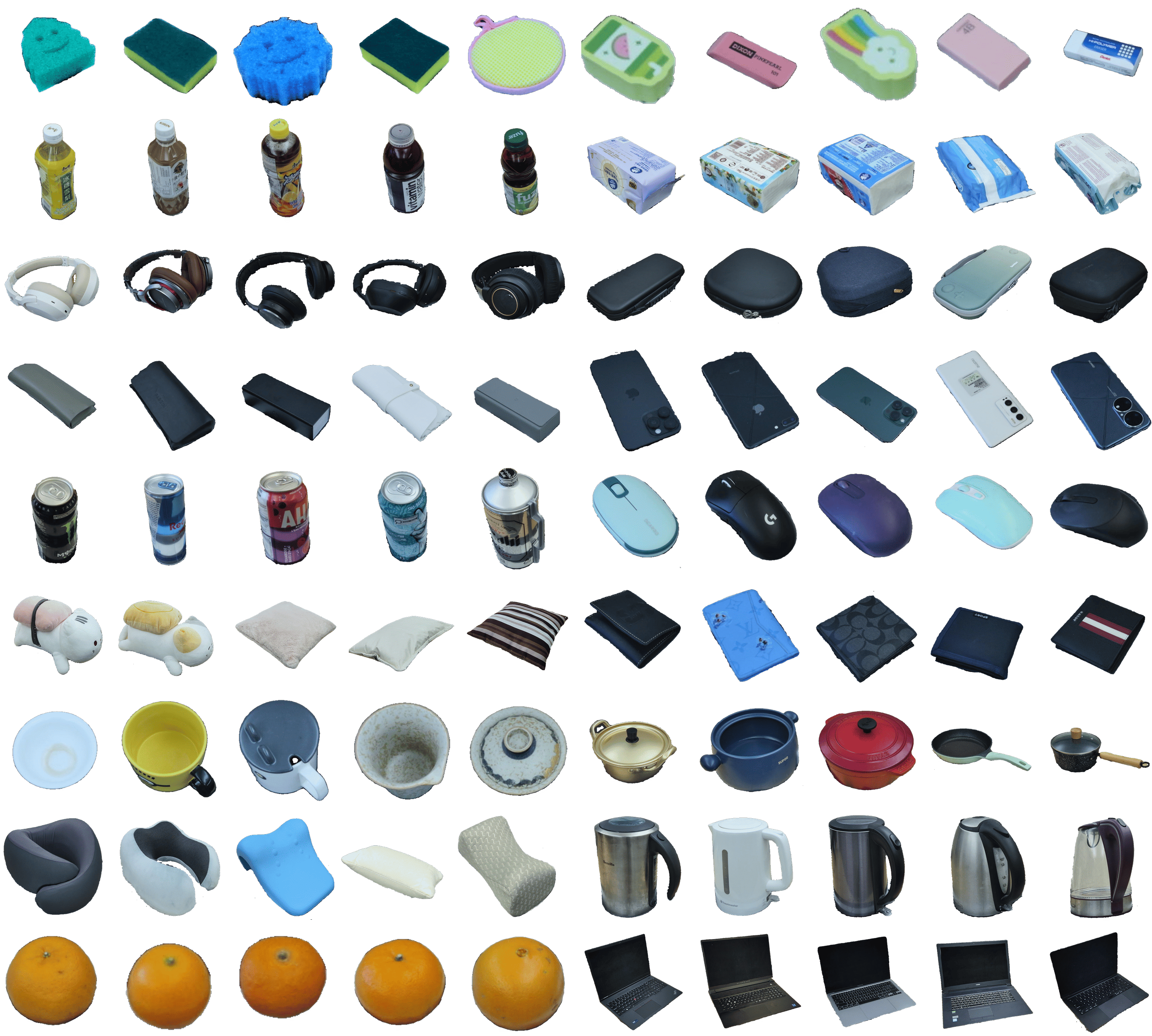}
\caption{RGB appearance of all 90 physical instances in \projectname{}, arranged by category with five instances per category. Thermal-state examples are shown in Fig.~\ref{fig:thermal_states}.}
\label{fig:inventory}
\end{figure}

\projectname{} includes 18 categories with five instances each: sponge, eraser, bottled drink, tissue, headset, storage bag, glasses case, smartphone, canned drink, mouse, pillow, wallet, cup, pot, cushion, electric kettle, orange, and laptop (Fig.~\ref{fig:inventory}). Every instance is captured at room temperature, about 22$^\circ$C. Instances in 11 categories are also captured in a second state produced by cooling, heating, or use, such as chilled cans, cups filled with hot water, and devices warmed by use. This gives 145 captured object states from 55 two-state and 35 one-state instances. Each capture is labeled with its instance and state. The label records how the capture was prepared, and the temperature values are in the data.

\subsection{Acquisition Setup and Protocol}

\begin{figure*}[t]
\centering
\includegraphics[width=1\linewidth]{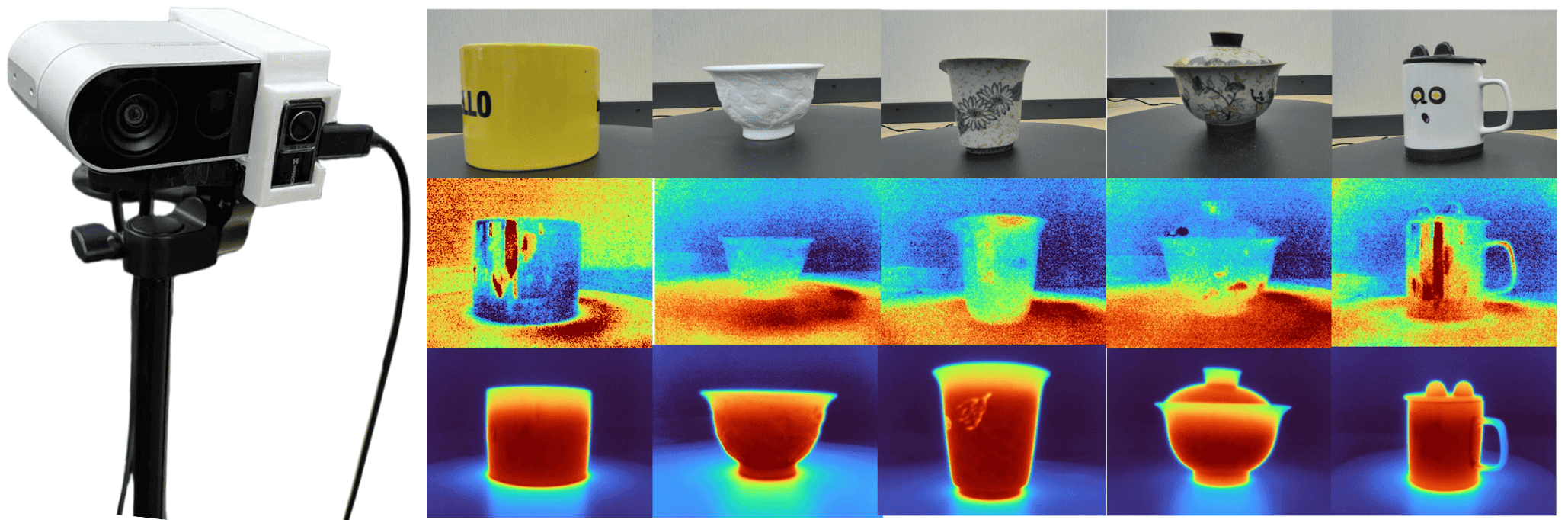}
\caption{RGB--thermal acquisition setup (left) and example RGB images and thermal visualizations (right). The two cameras are mounted on a shared bracket to collect paired views of each object.}
\label{fig:collection_setup}
\end{figure*}

The rig pairs the RGB stream of an ORBBEC Femto Bolt camera with a HIKMICRO Mini2 thermal camera on a rigid bracket, so their relative pose stays fixed (Fig.~\ref{fig:collection_setup}). The cameras are triggered together. Each view stores a 1920$\times$1080 RGB image, a 640$\times$576 depth map used later to recover the reconstruction scale (Section~\ref{sec:dataset_representation}), and the radiometric apparent temperature of every thermal pixel as a 256$\times$192 matrix in native camera coordinates.

For each object state, we acquire 10 evenly spaced views over a full turntable rotation at each of three camera elevations, giving 30 RGB--thermal pairs. A scan takes approximately five minutes. If the thermal view indicates cooling, we restore the heated state before continuing, for example by refilling hot water or operating the device again. The original images and temperature matrices are retained for reprocessing.

\subsection{RGB--Thermal Calibration and Pose Transfer}
\label{sec:calibration}

\begin{figure}[H]
\centering
\includegraphics[width=\linewidth]{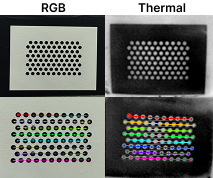}
\caption{The calibration target seen by the RGB camera (left) and the thermal camera (right). The top row shows the raw images of the white plate with 84 holes in front of the heated backing. The bottom row shows the detected hole centers, connected row by row in the same order in both images. These matched centers are the point correspondences used for the stereo calibration.}
\label{fig:calibration}
\end{figure}

\noindent\textbf{Calibration target.}
Visible-light targets gave weak or inconsistent features in our thermal images. For the stereo calibration we therefore use a white-painted wooden plate with an asymmetric grid of circular holes in front of a black-velvet backing warmed by an infrared lamp. The holes give visible contrast in RGB and temperature contrast in the thermal image, so both cameras detect the same circle centers (Fig.~\ref{fig:calibration}). Appendix~\ref{app:calib} gives the target geometry and detection details.

\noindent\textbf{Camera calibration.}
We use Zhang's calibration method~\citep{zhang1999calibration}. Thermal intrinsics and distortion parameters are estimated from 27 circle-grid frames, with all centers expressed at the native $256\times192$ resolution. RGB intrinsics are taken from the factory calibration. With the intrinsics fixed, stereo calibration on 24 circle-grid pairs estimates the rotation $R$ and metric translation $\mathbf t$ from the RGB camera to the thermal camera, with an RMS reprojection error of 0.3\,px. We denote this transform by $T_{\mathrm{th}\leftarrow\mathrm{rgb}}$, with $T_{a\leftarrow b}$ mapping coordinates from frame $b$ to frame $a$:
\begin{equation}
\mathbf X_{\mathrm{th}}=R\mathbf X_{\mathrm{rgb}}+\mathbf t.
\label{eq:extrinsic}
\end{equation}

\noindent\textbf{Camera-pose transfer.}
Thermal images in our setup provide limited texture for structure-from-motion pose estimation~\citep{schoenberger2016sfm,schoenberger2016mvs}. We therefore obtain per-view RGB camera-to-world poses $T^{(v)}_{\mathcal W\leftarrow\mathrm{rgb}}$ from the reconstruction and transfer them using the calibrated relative pose. The reconstruction poses and calibration use the same camera convention and metric scale:
\begin{equation}
T^{(v)}_{\mathcal W\leftarrow\mathrm{th}}
=T^{(v)}_{\mathcal W\leftarrow\mathrm{rgb}}
\left(T_{\mathrm{th}\leftarrow\mathrm{rgb}}\right)^{-1}.
\label{eq:posetransfer}
\end{equation}
We use the transferred pose to initialize each thermal view, then manually refine it by overlaying the projected mesh on the thermal image. The refined pose is stored with the view. This procedure uses calibrated camera geometry to register thermal observations to the reconstruction~\citep{lu2024thermalgaussian,zhang2024l2v2t2calib}.

\section{Dataset Representation and Derived Assets}
\label{sec:dataset_representation}

Each object instance includes the source measurements and the derived geometry and thermal assets.

\subsection{Object Geometry and RGB Textures}

We reconstruct each instance from its RGB views using MV-SAM3D~\citep{li2026mv} and export a GLB mesh with UV coordinates and a separate RGB texture. For instances with two thermal states, camera poses for all 60 views are estimated jointly, yielding a common reconstruction frame and one shared mesh. Metric scale is recovered by comparing rendered mesh depths with the recorded depth maps. Mesh coordinates and camera translations are expressed in meters. Using the captured object's geometry avoids the registration errors that can arise when mapping temperatures onto a CAD model with a different shape or scale. The shared geometry supports rendering both visible appearance and thermal state from the same object asset.

\subsection{Temperature Mapping and Thermal Textures}
\label{sec:registration}

For view $v$, let $\Phi^{(v)}$ be the raw apparent-temperature matrix and $K_{\mathrm{th}}$, $D_{\mathrm{th}}$ the thermal intrinsics and distortion. A mesh point $\mathbf X$ in the reconstruction frame, in homogeneous coordinates $\bar{\mathbf X}$, maps to a temperature by
\begin{align}
\mathbf x_{\mathrm{th}}^{(v)}
&=\left[\left(T^{(v)}_{\mathcal W\leftarrow\mathrm{th}}\right)^{-1}\bar{\mathbf X}\right]_{1:3},\\
\mathbf u^{(v)}
&=\pi\!\left(K_{\mathrm{th}},D_{\mathrm{th}},\mathbf x_{\mathrm{th}}^{(v)}\right),\\
\tau^{(v)}(\mathbf X)&=\Phi^{(v)}\!\left(\mathbf u^{(v)}\right),
\label{eq:projection}
\end{align}
where $\pi$ projects with the calibrated distortion model.

To construct the UV temperature map, we rasterize the mesh from each refined thermal-camera pose at the native 256$\times$192 resolution, using pixel centers undistorted with $D_{\mathrm{th}}$. A depth buffer retains the nearest surface, and back faces are culled. Each visible pixel contributes its original temperature value without interpolation. These values are splatted with bilinear weights into a 1024$\times$1024 UV map. Within each view, contributions are averaged by weight, and across views they are fused by their maximum:
\begin{equation}
T(q)=\max_{v}\;\frac{\sum_{p\in\mathcal V_v} w_{pq}\,\Phi^{(v)}[p]}{\sum_{p\in\mathcal V_v} w_{pq}},
\label{eq:fusion}
\end{equation}
where $\mathcal V_v$ is the set of visible pixels in view $v$ and $w_{pq}$ the bilinear weight of pixel $p$ at texel $q$. The weights only resample the image into UV space. The maximum keeps the highest reading where views disagree, so the map is an upper envelope of the observations rather than an average.

Unobserved texels are filled from the nearest observed texel within the same connected UV region. Regions without any observations remain marked as unobserved. We store the observed and filled temperature maps as floating-point arrays in degrees Celsius, alongside a validity mask distinguishing measured, filled, and unobserved texels. Thermal textures visualize the filled map using the Turbo color map over a fixed 10 to 60$^\circ$C range. This range applies only to visualization, and the original temperature matrices are retained for quantitative analysis. Figure~\ref{fig:thermal_states} compares thermal textures with the corresponding RGB reconstructions.

The stored values are apparent surface temperatures inferred from the detected radiation under the camera's settings. Geometric calibration and pose refinement do not correct for emissivity, reflections, or changes in ambient conditions.

\subsection{Data Organization and Statistics}

The dataset is organized by object instance and capture state (Table~\ref{tab:stats}). The 1,200 robot demonstrations described in Section~\ref{sec:evaluation} are a separate downstream dataset generated from these assets. Each instance includes camera calibration, refined per-view poses, a metric scale factor, and a manifest pairing the RGB images, thermal matrices, and depth maps. Appendix~\ref{app:assets} lists the file formats and metadata.

\begin{table}[t]
\centering\footnotesize
\caption{Object dataset overview.}
\label{tab:stats}
\begin{tabularx}{\linewidth}{@{}p{0.40\linewidth}Y@{}}
\toprule
Object categories & 18 \\
Physical instances & 90 (5 per category) \\
Thermal states & Room temperature for all 90 instances, plus one cooled, heated, or post-use state for 55 instances in 11 categories (145 object states) \\
Views per object state & 30 (3 elevations $\times$ 10 azimuths) \\
RGB--thermal pairs & 4,350 (145 object states $\times$ 30 views) \\
Raw data per view & RGB image (1920$\times$1080), apparent-temperature matrix (256$\times$192, CSV), depth map (640$\times$576) \\
Derived assets per instance & Mesh (GLB with UV), RGB texture, temperature UV map with validity mask, thermal texture (PNG), per-view camera poses, calibration \\
\bottomrule
\end{tabularx}
\end{table}
\section{Evaluation}
\label{sec:evaluation}

We construct three tasks to evaluate temperature-related instruction following in simulation. \textbf{H1} and \textbf{H2} require selecting a hot or iced object, or its room-temperature counterpart, from visually identical candidates. \textbf{H3} requires selecting between object categories whose candidates share the same thermal-state label.

\begin{table}[h]
\centering
\footnotesize
\setlength{\tabcolsep}{3pt}
\renewcommand{\arraystretch}{1.08}
\caption{Task configurations and complementary instructions.}
\label{tab:tasks}

\begin{tabularx}{\columnwidth}{
    @{}l l >{\raggedright\arraybackslash}X@{}
}
\toprule
Task / Scene & Field & Configuration \\
\midrule

H1 / Study
& Candidates & Identical white cups: hot / room temperature \\
& Distractors & Hot kettle; room-temperature eraser \\
& Targets & Hot cup / normal cup \\

\addlinespace[3pt]
H2 / Kitchen
& Candidates & Identical cans: iced / room temperature \\
& Distractors & Iced orange; room-temperature cup \\
& Targets & Iced can / normal can \\

\addlinespace[3pt]
H3 / Kitchen
& Candidates & Phone / sponge, both hot \\
& Distractors & Room-temperature orange; hot cup \\
& Targets & Hot phone / hot sponge \\

\bottomrule
\end{tabularx}

\par\vspace{3pt}
{\scriptsize\raggedright
Each instruction is ``pick up the [target] and place it
on the mat.'' Targets are listed in the order of
Instructions 1 and 2.
``Normal'' denotes room temperature.
\par}
\end{table}

\subsection{Task Suite and Scene Construction}

\begin{figure*}[t]
\centering
\includegraphics[width=\linewidth]{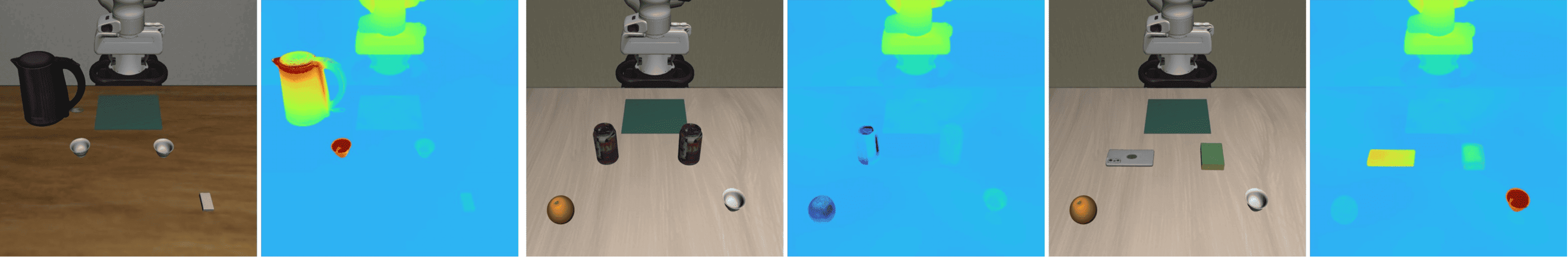}
\caption{LIBERO-derived task configurations H1--H3. Each task is shown through aligned front RGB and thermal views. The policy additionally receives a wrist-mounted RGB view. The mat specifies the placement destination; additional objects serve as distractors.}
\label{fig:simulation_scenes}
\end{figure*}

We construct three pick-and-place tasks using \projectname{} assets in LIBERO-derived environments~\citep{liu2023libero}. H1 uses a study-table scene, while H2 and H3 use a kitchen-table scene. All tasks use the same Panda arm, parallel-jaw gripper, and operational-space pose controller. Each instruction specifies one of two candidate objects and asks the robot to place it on a fixed tabletop mat. As \projectname{} contains a different set of objects from LIBERO, we replace the original scene objects with visual and thermal assets from our captured objects, using them as manipulation targets, background objects, and distractors. Table~\ref{tab:tasks} lists the candidates, distractors, and instruction pairs. Fig.~\ref{fig:simulation_scenes} shows the three scenes.

In H1 and H2, the candidates share geometry, RGB texture, scale, mass, and friction. Thermal input provides the cue for distinguishing their states. H3 evaluates category selection between a phone and a sponge, both labeled hot. Their geometry and thermal images differ, so H3 measures performance on category-based instructions rather than isolating recognition from RGB input.

Each layout supports both instructions with identical initial object and robot states. Candidate identities or states are assigned evenly to the two placement slots. Positions are sampled within a $5\times5$\,cm square around each nominal location, while orientations remain fixed. Distractors receive the same bounded positional perturbation around peripheral locations, with visibility checked during collection. The mat, robot and gripper initialization, camera poses, image orientation, and lighting remain fixed. Physical parameters are fixed per asset across episodes, while different object categories retain their own geometry and simulation-scale parameters.

Thermal views are rendered from object-attached pseudo-color textures using the front RGB camera pose. The textures move with the objects, but their temperature fields remain fixed throughout the rollouts. The simulator does not model heat transfer or changes due to contact, heating, or cooling.

\subsection{Data Collection}

For each task, we collect demonstrations for both instructions in 200 training layouts, yielding 400 demonstrations per task and 1,200 in total. Each candidate-to-slot assignment appears in 100 layouts per task. A scripted expert uses simulator state to approach, grasp, lift, transport, and release the target. Collection checks verify the simulator's placement goal and candidate visibility. Additional manipulation diagnostics are stored with the trajectories.

The observations comprise front RGB, front thermal, wrist RGB, robot state, and the instruction. The expert's privileged object poses and target identifiers are used for collection and auditing, but are not included in the policy observation. All 1,200 demonstrations are used for optimization.

\subsection{Policy Training}

Both conditions start independently from the same pretrained $\pi_{0.5}$ LIBERO checkpoint~\citep{pi05}. The thermal condition receives the instruction $\ell$, robot state $\mathbf{s}_t$, front RGB image $I_t^{\mathrm f}$, wrist RGB image $I_t^{\mathrm w}$, and aligned front thermal image $I_t^{\mathrm{th}}$:
\begin{equation}
\hat{\mathbf a}_{t:t+H-1}
=\pi_\theta\!\left(I_t^{\mathrm f},I_t^{\mathrm w},I_t^{\mathrm{th}},\mathbf{s}_t,\ell\right).
\label{eq:policy}
\end{equation}
The thermal view occupies the otherwise masked third image slot of the single-arm interface. The shared vision encoder processes each view separately. The policy receives rendered pseudo-color thermal images, not numerical temperature matrices, and uses no additional thermal-specific pretraining. Each action comprises a six-dimensional operational-space pose increment and a gripper command.

We fine-tune all model parameters, including the vision encoder, language backbone, and action expert, using the action flow-matching objective. In the RGB-only condition, the thermal image is zeroed and its image mask is disabled during both training and inference. Both conditions otherwise use the same demonstrations, instructions, actions, normalization statistics, base checkpoint, and optimization settings.

We compare the two policies at 10,000 optimizer updates. Training uses NVIDIA H100 hardware with a global batch size of 512 and seed 42.

\subsection{Results}
Both policies are evaluated on the same 60 episodes: 10 per instruction and 20 per task. We report overall success and per-task success to distinguish thermal-state selection from category selection.

Table~\ref{tab:evaluation_results} compares the policies after 10,000 optimization updates. The RGB--thermal policy succeeds in 49 of 60 episodes (81.7\%), compared with 34 of 60 (56.7\%) for the RGB-only baseline, a gain of 25.0 percentage points.

The gains occur on H1 and H2, where the candidate objects are visually identical but have different thermal states. Success rises from 35.0\% to 75.0\% on each task, an increase of 40.0 percentage points. All four instructions improve, including those requesting the room-temperature counterpart of a hot or iced object. Adding thermal input therefore enables the VLA to follow temperature-related language instructions more successfully than the RGB-only baseline. 

Inspection of the failed evaluation videos for H1 and H2 showed that the robot selected the correct target in every inspected episode (100\%). This supports the VLA's ability to use thermal information to identify the object specified by a temperature-related instruction. These episodes nevertheless failed because execution exceeded the time limit: the robot either did not close the gripper or transported the grasped object to the mat without releasing it. Such failures may reflect difficulty learning the timing of grasping and release, together with limited recovery behavior when execution deviates from the scripted demonstrations. 

On H3, the RGB--thermal policy succeeds in 19 of 20 episodes (95.0\%), compared with 20 of 20 (100.0\%) for the RGB-only baseline. It thus retains high success when selecting between object categories, alongside the gains on H1 and H2. The observed difference is small in this task and does not establish whether thermal input affects visual recognition more generally.

Despite these gains, thermal-state selection on H1 and H2 remains less reliable than category selection on H3. $\pi_{0.5}$ inherits its visual representations from a pretrained VLM~\citep{pi05}. One possible explanation for this gap is limited exposure to thermal imagery during VLM pretraining, leaving these representations less adapted to thermal patterns than to RGB appearance. Our fine-tuning may therefore be insufficient to fully learn the association between thermal patterns and temperature-related instructions. Broader thermal-language training data could improve this ability. A controlled study of pretraining coverage would be needed to test this explanation.

Together, these experiments show that \projectname{} assets can add temperature-dependent observations and instructions to LIBERO-derived manipulation tasks. The resulting demonstrations support fine-tuning thermal-VLA policies that perform better than an RGB-only baseline on thermal-state selection.

\begin{table}[t]
\centering
\footnotesize
\setlength{\tabcolsep}{4pt}
\renewcommand{\arraystretch}{1.1}
\caption{Task success rates (\%) with and without thermal observations.}
\label{tab:evaluation_results}

\begin{tabular}{@{}llccc@{}}
\toprule
Task & Thermal & Instr.~1 & Instr.~2 & Average \\
\midrule

\multirow{2}{*}{H1}
& w/  & 80.0 & 70.0 & 75.0 \\
& w/o & 30.0 & 40.0 & 35.0 \\

\addlinespace[3pt]
\multirow{2}{*}{H2}
& w/  & 70.0 & 80.0 & 75.0 \\
& w/o & 40.0 & 30.0 & 35.0 \\

\addlinespace[3pt]
\multirow{2}{*}{H3}
& w/  & 90.0  & 100.0 & 95.0 \\
& w/o & 100.0 & 100.0 & 100.0 \\

\midrule
\multirow{2}{*}{Overall}
& w/  & -- & -- & \textbf{81.7} \\
& w/o & -- & -- & \textbf{56.7} \\
\bottomrule
\end{tabular}

\par\vspace{3pt}
{\scriptsize\raggedright
Instructions 1 and 2 follow the order in
Table~\ref{tab:tasks}, with 10 episodes per instruction
per policy. Overall averages give equal weight to the
three tasks (H1--H3), totaling 60 episodes per policy.
\par}
\end{table}

\section{Conclusion}
\label{sec:conclusion}

We presented \projectname{}, a dataset of 90 physical objects from 18 categories that associates measured apparent surface temperatures with object appearance and reconstructed geometry. The dataset provides raw measurements and aligned assets for simulation. In three LIBERO-derived tasks, adding rendered thermal observations to a fine-tuned VLA increases overall success from 56.7\% to 81.7\%, with gains on temperature-related instructions and high success on category selection. \projectname{} provides a basis for constructing simulation tasks and training policies that use object temperature to interpret and execute language instructions.

\section{Limitations and Future Work}
\label{sec:limitations}

\noindent\textbf{Dynamic thermal simulation.}
The current assets encode static apparent surface temperatures and do not model heat transfer during interaction. One direction is to couple rigid-body simulation with a reduced-order thermal model, representing object regions by their heat capacities and connections by thermal conductances. Such a model could update surface temperatures and rendered observations as objects exchange heat with each other and the environment. Developing and validating it would require material annotations and time-resolved measurements of heating, cooling, and contact.

\noindent\textbf{Automated acquisition and scene expansion.}
Thermal-view poses require manual refinement after calibrated pose transfer, which limits collection throughput. We plan to investigate ways to reduce this manual effort and automate more of the acquisition and reconstruction pipeline. We also aim to include larger objects and construct scene-level RGB--thermal--3D environments.

\noindent\textbf{Thermal-VLA manipulation.}
The current evaluation covers three simulation tasks, with the same object identities in training and evaluation. Future work could use spatial temperature distributions to guide grasp selection and action sequences, such as grasping a hot wok by a cooler handle or placing an insulating trivet before setting down a hot pot. Evaluating these policies on unseen objects and real robots would test whether the learned behavior transfers beyond the current simulation setting.

\useRomanappendicesfalse
\appendices

\section{Calibration Target and Projection Details}
\label{app:calib}

\subsection{Target Geometry and Feature Detection}

The white-painted wooden target measures $200\times150$\,mm, with 8\,mm holes at 12\,mm horizontal spacing and a 10.3923\,mm interlaced vertical offset. Eight rows alternate between 11 and 10 holes, for 84 centers. A black-velvet backing behind the plate is warmed by an infrared lamp, and the same hole centers serve both cameras.

RGB circle centers are detected directly as dark blobs in the grayscale image. Each thermal temperature matrix is clipped at per-frame percentile bounds, scaled to 8-bit grayscale, and upsampled four times before blob detection. Each center is then refined as an intensity-weighted centroid and rescaled to the native $256\times192$ resolution. The 84 centers are ordered by the known row structure of the target, so both cameras index the same hole. The same thermal detections serve the thermal intrinsic calibration and the stereo calibration. Enhancement serves detection only and never alters the raw matrices. All calibration uses OpenCV with a five-parameter distortion model.

\subsection{Transform Convention}

$T_{\mathrm{th}\leftarrow\mathrm{rgb}}$ is the homogeneous form of Eq.~\eqref{eq:extrinsic}, and its inverse has rotation $R^\top$ and translation $-R^\top\mathbf t$. Eq.~\eqref{eq:posetransfer} needs this inverse because the reconstruction provides camera-to-world poses. After scale recovery (Section~\ref{sec:dataset_representation}), reconstruction coordinates and the calibration translation are both in meters. The stored thermal matrices are kept in the native, unrectified sensor coordinates. Projection uses the pinhole model with the five-parameter distortion $D_{\mathrm{th}}$, applied by undistorting pixel centers before rasterization, so the stored measurements are never resampled. Circle centers detected on an upsampled image for the stereo calibration are rescaled to the native 256$\times$192 resolution first.

\section{Data Files and Asset Provenance}
\label{app:assets}

The release is organized by object instance. The source data per view are the RGB image (JPEG, 1920$\times$1080), its object cutout (PNG), the depth map (16-bit, millimeters, 640$\times$576) with a per-frame JSON file of camera intrinsics, and the apparent-temperature matrix (CSV, 256$\times$192, degrees Celsius). A manifest pairs these files by acquisition order. The calibration files hold the RGB and thermal intrinsics with distortion coefficients, the extrinsics of Eq.~\eqref{eq:extrinsic}, the per-view RGB poses from the reconstruction, the refined per-view thermal poses, and the recovered scale factor. The derived files per instance are the mesh (GLB with UV coordinates), the RGB texture (PNG), and, per state, the observed and filled temperature maps (floating-point, degrees Celsius, 1024$\times$1024 in UV layout), a validity mask marking observed, filled, and unobserved texels, the Turbo texture (PNG), and a provenance file listing the source views. Mesh coordinates and camera translations are expressed in meters. Camera poses are camera-to-world transforms in the OpenCV convention. Unobserved texels are identified through the mask rather than a sentinel value, and every derived file names its instance, state, and source views. Source measurements remain distinct from pseudo-color renderings.

The reconstruction pipeline uses MV-SAM3D~\citep{li2026mv}, and where SAM 3D Objects code is used its license applies~\citep{sam3dobjectslicense}. Calibration and projection use OpenCV. Camera vendor software is not redistributed. The downstream study depends on LIBERO and the openpi implementation of $\pi_{0.5}$, starting from its released LIBERO checkpoint. Exact software versions are listed in the release documentation.

\bibliographystyle{IEEEtranN}
\bibliography{references}
\end{document}